\documentclass[10pt,twocolumn,letterpaper]{article}

\usepackage{cvpr}              

\usepackage{multirow}
\usepackage{makecell}
\usepackage{graphicx}
\usepackage{amsmath}
\usepackage{amssymb}
\usepackage{booktabs}
\usepackage{url}

\usepackage[pagebackref,breaklinks,colorlinks]{hyperref}

\usepackage[capitalize]{cleveref}
\crefname{section}{Sec.}{Secs.}
\Crefname{section}{Section}{Sections}
\Crefname{table}{Table}{Tables}
\crefname{table}{Tab.}{Tabs.}

\def\cvprPaperID{4068} 
\def\confName{CVPR}
\def\confYear{2022}

\begin{document}

\title{VISTA: Boosting 3D Object Detection via Dual Cross-VIew SpaTial Attention}

\author{Shengheng Deng\textsuperscript{1,*},
Zhihao Liang\textsuperscript{1,3,*},
Lin Sun\textsuperscript{2} and Kui Jia\textsuperscript{1,4,\textdagger}\\
\textsuperscript{1}South China University of Technology,\quad \textsuperscript{2}Magic Leap, Sunnyvale, CA \\
\textsuperscript{3}DexForce Technology Co., Ltd.,\quad \textsuperscript{4}Peng Cheng Laboratory\\
{\tt\small  \{eedsh, eezhihaoliang\}@mail.scut.edu.cn},
{\tt\small kuijia@scut.edu.cn},
{\tt\small lsun@magicleap.com}
}

\maketitle

\let\thefootnote\relax\footnote{* indicates equal contribution.}

\let\thefootnote\relax\footnote{\textsuperscript{\textdagger}Correspondence to Kui Jia $<$kuijia@scut.edu.cn$>$.}

\begin{abstract}
   Detecting objects from LiDAR point clouds is of tremendous significance in autonomous driving. In spite of good progress, accurate and reliable 3D detection is yet to be achieved due to the sparsity and irregularity of LiDAR point clouds. Among existing strategies, multi-view methods have shown great promise by leveraging the more comprehensive information from both bird's eye view (BEV) and range view (RV). These multi-view methods either refine the proposals predicted from single view via fused features, or fuse the features without considering the global spatial context; their performance is limited consequently. In this paper, we propose to adaptively fuse multi-view features in a global spatial context via Dual Cross-VIew SpaTial Attention (VISTA). The proposed VISTA is a novel plug-and-play fusion module, wherein the multi-layer perceptron widely adopted in standard attention modules is replaced with a convolutional one. Thanks to the learned attention mechanism, VISTA can produce fused features of high quality for prediction of proposals. We decouple the classification and regression tasks in VISTA, and an additional constraint of attention variance is applied that enables the attention module to focus on specific targets instead of generic points. We conduct thorough experiments on the benchmarks of nuScenes and Waymo; results confirm the efficacy of our designs. At the time of submission, our method achieves 63.0\% in overall mAP and 69.8\% in NDS on the nuScenes benchmark, outperforming all published methods by up to 24\% in safety-crucial categories such as cyclist. 
   \href{https://github.com/Gorilla-Lab-SCUT/VISTA}{Code}.
\end{abstract}

\section{Introduction}
\label{sec:intro}

\begin{figure}[!htb]
    \centering
    \includegraphics[width=1.0\linewidth]{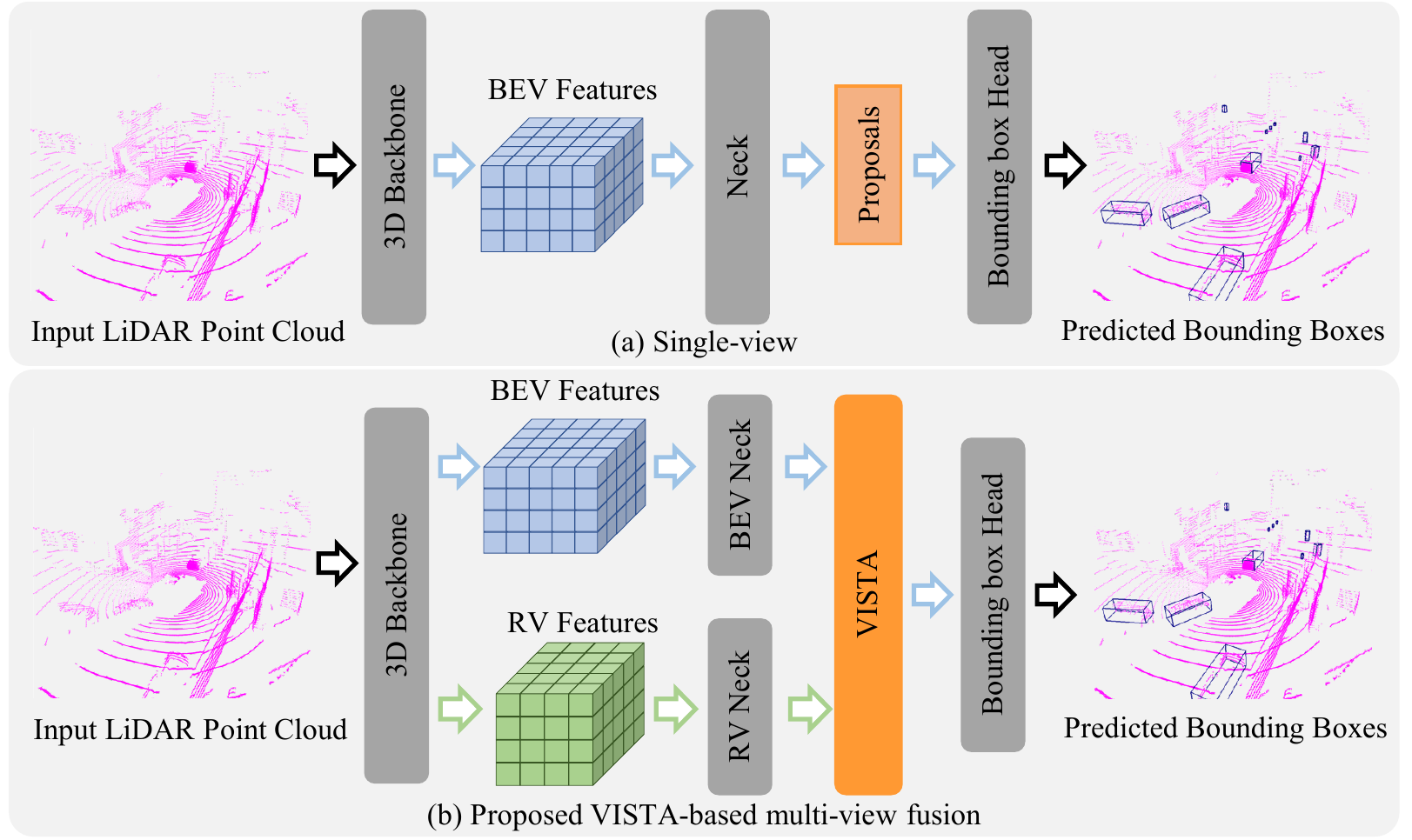}
    \vspace{-0.7cm}
    \caption{Comparison between the single-view detection and the proposed VISTA-based multi-view fusion. (a) shows the single-view detection pipeline. (b) illustrates the proposed VISTA-based multi-view fusion. The BEV and RV features which are extracted by a shared 3D backbone are passed into individual necks, and pass through the VISTA to output high quality fused features.}
    \label{figPipelines}
    \vspace{-0.5cm}
\end{figure}


LiDAR is one of the prominent sensors which is widely used in autonomous driving to provide precise 3D information of the objects. Therefore, LiDAR based 3D object detection has attracted a lot of attention. Many 3D object detection algorithms \cite{zhou2018voxelnet,yang2018pixor,lang2019pointpillars} apply the convolutional neural networks to 3D point clouds by voxelizing the unordered and irregular point clouds into volumetric grids. Nevertheless, the 3D convolutional operator is computationally inefficient and memory-consuming. To mitigate these issues, a line of works \cite{chen2020object,yan2018second} utilize sparse 3D convolutions \cite{graham2014spatially,graham2017submanifold,Graham_2018_CVPR} in the network backbones to extract features. As illustrated in the Figure \ref{figPipelines} (a), these works project the 3D feature maps into the bird's eye view (BEV) or range view (RV), and the object proposals are produced from these 2D feature maps using proposal methods.

Different views have their own advantages and drawbacks to consider. In BEV, objects do not overlap with each other and the size of each object is invariant to the distance from the ego-vehicle. RV is the native representation of LiDAR point clouds, therefore, it can produce compact and dense features. However, projection would inevitably impair the integrity of spatial information conveyed in the 3D space no matter which of BEV or RV is chosen. For example, due to the self-occlusion and the characteristic of LiDAR data generation, the BEV representation is extremely sparse and it consolidates the height information of 3D point clouds, occlusion and variation in object size would be severer in RV since it loses depth information. Obviously, learning jointly from multiple views, a.k.a multi-view fusion provides us a solution to accurate 3D object detection. Some of the previous multi-view fusion algorithms \cite{chen2017multi,ku2018joint} produce the proposals from a single view and utilize the multi-view features to refine proposals. Performances of such algorithms highly depend on the quality of the produced proposals; however, proposals generated from a single view make no use of all the available information, possibly leading to suboptimal solutions. Other works \cite{chen2020every,zhou2020end} fuse the multi-view features according to the coordinate projection between different views. Accuracy of such fusion methods relies on the complementary information provided in the corresponding region of the other view; yet the occlusion effect is inevitable, inducing low-quality multi-view feature fusion.

To boost the performances of 3D object detection, in this paper, given learned 3D feature maps from both BEV and RV, we propose to produce high quality fused multi-view features from the global spatial context via Dual Cross-VIew SpaTial Attention (VISTA) for proposal prediction, as demonstrated in Figure \ref{figPipelines} (b). The proposed VISTA utilizes the attention mechanism originated in the transformer which is successfully applied to various research context (e.g. natural language processing, 2D computer vision). Compared with direct fusion via coordinate projections, the inbuilt attention mechanism in VISTA exploits the global information and adaptively models all the pairwise correlations across views by treating features of individual views as sequences of feature elements. To model the cross-view correlations comprehensively, the local context in both views must be taken into account, thus we replace the MLPs in the conventional attention module with the convolutional operators, of which we show the effectiveness in the Section \ref{sec:exp}. Nevertheless, learning the correlations across views is still challenging, as shown in Section \ref{sec:exp}. Directly adopting the attention mechanism for multi-view fusion brings little gains and thus, we argue that it is mainly due to the characteristic of the task 3D object detection itself.

Generally, the 3D object detection task could be divided into two sub-tasks: classification and regression. As elaborated in \cite{meyer2019lasernet,chen2020every}, the 3D object detector faces many challenges when detecting objects in the whole 3D scenes, such as occlusion, background noise and the scarce texture information of point cloud. In consequence, the attention correlations are difficult to learn and the attention module tends to learn the mean of the whole scene, which is unexpected as the attention module is designed for paying attention to regions of interest. Therefore, we explicitly constrain the variance of the attention maps learned by the attention mechanism, which guides the attention module to be aware of the meaningful regions in the complex 3D outdoor scenes. Moreover, different learning targets for classification and regression determine the different expectations of the learned queries and keys in the attention module. The various regression targets (e.g. scale, translate) across different objects expect the queries and keys to be aware of the characteristic of the objects. The classification task instead, pushes the network to understand the common properties of the object classes. Inevitably, sharing the same attention modeling will bring conflicts into the training of these two tasks. Furthermore, on one hand, due to the loss of texture information, it is difficult for neural networks to extract semantic features from point clouds. On the other hand, the neural networks can easily learn the geometric property of objects from point clouds. As a result, during training, a dilemma that the classification being dominant by the regression is aroused. To tackle these challenges, we decouple these two tasks in the proposed VISTA to learn to aggregate different cues in terms of different tasks. 

Our proposed VISTA is a plug-and-play module and can be adopted to the recent advanced target assign strategies. We test our proposed VISTA-based multi-view fusion on different target assign algorithms on the benchmark datasets of nuScenes \cite{caesar2020nuscenes} and Waymo \cite{sun2020scalability}. Ablation studies on their validation sets confirm our conjecture. Thanks to the high quality fused features produced by the proposed VISTA, our proposed method outperforms all the published algorithms. At the time of submission, our final results achieve 63.0\% in overall mAP and 69.8\% in NDS on nuScenes leaderboard. On Waymo Open Dataset, we achieve 74.0\%, 72.5\%, and 71.6\% level 2 mAPH on vehicle, pedestrian and cyclist. We summarize our main contributions as follows. 

\begin{itemize}

\item We propose a novel plug-and-play fusion module Dual Cross-VIew SpaTial Attention (VISTA) to produce well-fused multi-view features to boost the performances of 3D object detector. Our proposed VISTA replaces the MLPs with convolutional operators, which is capable of better handling the local cues for attention modeling.

\item We decouple the regression and classification tasks in the VISTA to leverage individual attention modeling to balance the learning of these two tasks. We apply the attention variance constraint to VISTA during training phase, which facilitate the attention learning and empower the network to attend to the regions of interest.

\item We conduct thorough experiments on the benchmark datasets of nuScenes and Waymo. Our proposed VISTA-based multi-view fusion can be adopted in various advanced target assign strategies, easily boost the original algorithms and achieve state-of-the-art performances on the benchmark datasets. Specifically, our proposed method outperforms the second best methods by 4.5\% in overall performance, and up to 24\% on the safety-crucial object categories like cyclist.
\end{itemize} 

\section{Related Works}

\subsection{Single-View 3D Detection}

\textbf{BEV-Based 3D Detection} Most of the voxel-based 3D detection algorithms detect objects on BEV. \cite{lang2019pointpillars} projects the point clouds into BEV pillars, and feeds the BEV pillars into 2D CNNs. Such a projection inevitably induces 3D spatial information loss. Recent works \cite{zhou2018voxelnet,chen2020object,yan2018second,yin2020center} mitigate this issue by utilizing 3D CNNs to directly operate on 3D point clouds or processed 3D point clouds, e.g. voxels, and then project 3D feature maps into BEV, and finally detect objects on projected BEV features.

\textbf{RV-Based 3D Detection} Few works \cite{meyer2019lasernet} detect objects from RV. RV representations provide compact features. However, as mentioned in \cite{meyer2019lasernet}, the RV detectors need more training data, and there exist great challenges posed by occlusion and variant object scales with range.

Nevertheless, both projections will damage the 3D spatial information integrity. We believe that a comprehensive 3D detection framework needs to learn from both views, and the performance is decided by the fused features with the complementary information from both views.

\subsection{Multi-View 3D Detection}\label{subsubsecMultiView}

A line of works \cite{chen2020mvlidarnet, laddha2021mvfusenet} realize multi-view fusion either by aggregating features to refine proposals or fusing features in the region constrained by the spatial projection. \cite{chen2017multi, ku2018joint} fuse the ROI features from point cloud and camera image for proposals refinement. Instead of fusing multi-view features at the ROI level, \cite{zhou2020end} fuses point-wise features from BEV and RV. Different from previous works, CVCNet \cite{chen2020every} proposes a hybrid voxelization method to unify the benefits from both views, and utilizes hough transform to restrict the consistency between classification results from both views. However, the CVCNet does not utilize the multi-view features to produce proposals directly, thus fails to make full use of the fused features to do the 3D detection. 

All of these works fuse features in limited regions or do not exploit the fused feature for the 3D detection. To leverage the multi-view features from the global spatial context, our proposed VISTA considers the interactions among features from different views in the entire scene.

\subsection{Attention in Transformer}

Thanks to the capability of effectively capturing long-range dependencies of features in the input feature sequences, transformer \cite{vaswani2017attention} have been widely transferred into computer vision task \cite{dosovitskiy2020image,han2021transformer,liu2021swin, carion2020end,zhu2020deformable, guo2021pct, sheng2021improving}. The core component in transformer is the self-attention module, which explicitly models the pair-wise correlations among the input feature sequences. ViT \cite{dosovitskiy2020image} divides the images into patches for attention construction to realize image classification. PCT \cite{guo2021pct} modifies the attention module into a discrete Laplacian operator for point cloud classification. CT3D \cite{sheng2021improving} reweights the proposal features via channel-wise attention for bounding box refinement.

Unfortunately, due to the characteristic of 3D object detection task itself and the inherent property of outdoor 3D point clouds, the existing attention module fails to focus on the regions of the interest in the scenes. Moreover, the network training will be easily dominant by the regression task. The proposed VISTA, instead, addresses the above issues via the decoupled attention modeling and the designed training constraint, thus is able to produce high quality fused multi-view features for 3D object detection.

\section{Overview} 

\begin{figure*}[!tb]
    \centering
    \includegraphics[width=1.0\textwidth]{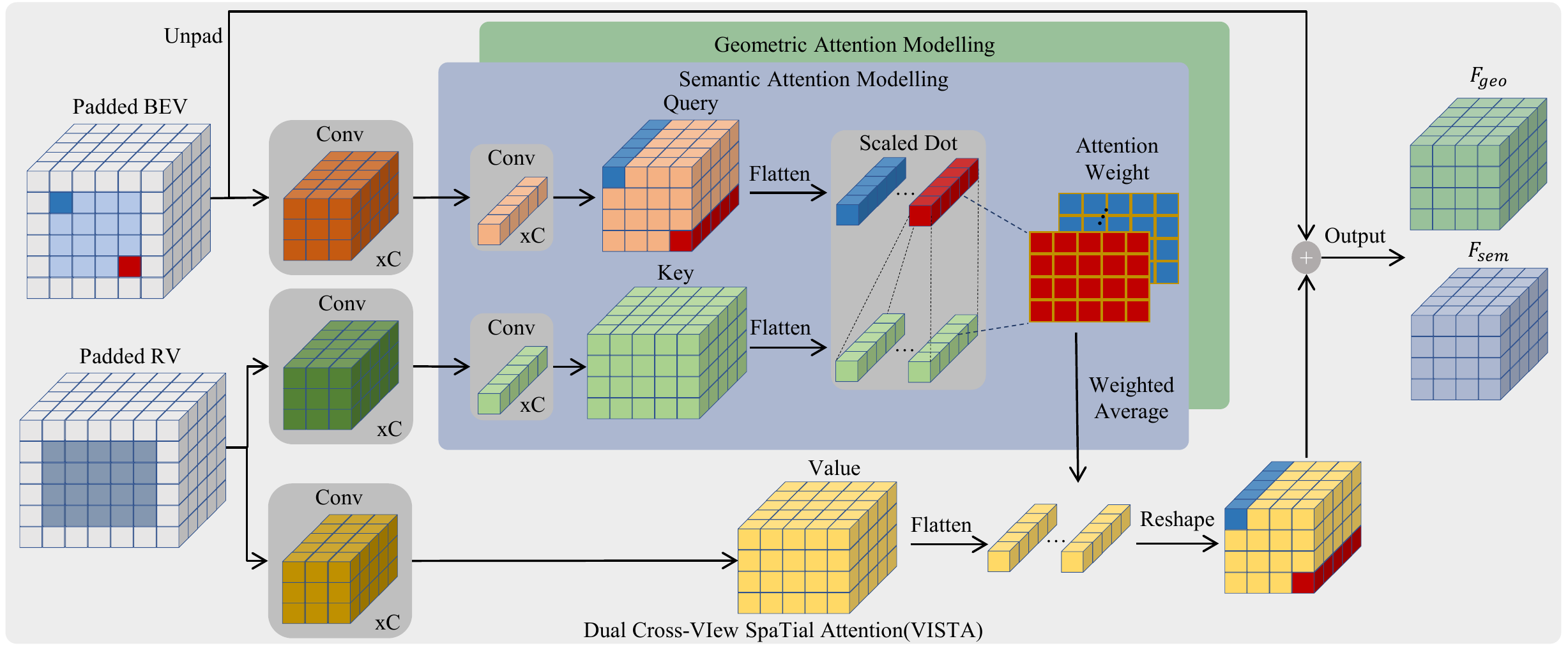}
    \vspace{-0.8cm}
    \caption{The architecture of the proposed VISTA.}
    \label{figArchitecture}
    \vspace{-0.5cm}
\end{figure*}

Given LiDAR point clouds from one scene, the task of 3D object detection is to accurately predict the categories of objects and output the oriented bounding boxes enclosed the objects in the scene. As illustrated in Figure \ref{figPipelines} (a), in most 3D object detectors, the learned 3D feature grids will be collapsed into 2D feature maps of BEV or RV, followed by a 2D bounding box head. We term these methods as single-view detector. Inevitably, the 'collapse' operation will impair the spatial integrity, which will lead to inferior bounding box prediction. To compensate the loss, various multi-view fusion methods have been proposed. As elaborated in Section \ref{sec:intro}, hindered by the quality of proposals or the limited region of fusion, all of these methods cannot fuse multi-view features comprehensively. We argue that one should consider the global context during multi-view fusion to better leverage the complementary information conveyed in two views. Therefore, we propose Dual Cross-VIew SpaTial Attention (VISTA) module which decouples the classification and regression tasks in 3D object detection, and is trained under designed attention constraint.

We follow the general fusion pipeline that is widely used in existing fusion algorithms. The overall architecture is demonstrated in Figure \ref{figPipelines} (b). We adopt the widely used sparse 3D ResNet \cite{chen2020every} as our shared 3D backbone to produce 3D feature maps $F_{3d}\in \mathbb{R}^{B\times C\times H\times W\times D}$, where the B is the batch size, C is the feature dimension, H, W, and D are the size of $F_{3d}$ corresponding to width, height, and depth axes, respectively. For BEV and RV, $F_{3d}$ is collapsed into two 2D feature maps $F_{bev}\in \mathbb{R}^{B\times (C\times W)\times H \times D}$ and $F_{rv}\in \mathbb{R}^{B\times (C\times D)\times H \times W}$. $F_{bev}$ and $F_{rv}$ are fed into individual 2D feature extractor, namely 2D neck. As most recent state-of-the-art detectors \cite{yin2020center,chen2020object}, we adopt UNet-like architecture as our 2D neck which contains several convolutional layers and each of which is followed by a normalization and activation function. After 2D necks, the VISTA takes $F_{bev}$ and $F_{rv}$ as inputs, and outputs the fused multi-view features for producing the detection results.


\section{Dual Cross-VIew SpaTial Attention}

For most voxel-based 3D detectors that densely produce pillar-wise proposals, generating nutrient feature maps guarantees the detection qualities empirically. In the case of multi-view 3D object detection where the proposal comes from the fused feature maps, an overall consideration of global spatial context during fusion is required. To this end, we seek to utilize the ability of capturing global dependencies of attention module for multi-view fusion, namely cross-view spatial attention. Before considering the global context, the cross-view spatial attention module needs to aggregate the local cues for constructing the correlations between different views, as we shown in Section \ref{sec:exp}. Therefore, we are motivated to propose VISTA, wherein the standard attention module based on multi-layer perceptron is replaced with a convolutional one. However, learning the attention in the complicated 3D scenes is difficult. To adopt the cross-view attention for multi-view fusion, we further decouple the classification and regression tasks in the VISTA, and apply the proposed attention constraint to boost the learning process of attention mechanism.

In this section, we will first introduce the overall architecture of the proposed Dual Cross-VIew SpaTial Attention (VISTA) in detail, then we elaborate the decoupling design and the attention constraints for the proposed VISTA.

\subsection{Overall Architecture}

As shown in Figure \ref{figArchitecture}, VISTA takes feature sequences from two different views as inputs and models the cross-view correlations among the multi-view features. Unlike the vanilla attention module that uses linear projections to transform input feature sequences, VISTA projects the input feature sequences $\mathcal{X}_1\in \mathbb{R}^{n\times d_f}$ and $\mathcal{X}_2\in \mathbb{R}^{m\times d_f}$ into queries $\mathcal{Q}\in \mathbb{R}^{n\times d_q}$ and keys $\mathcal{K}\in \mathbb{R}^{m\times d_q}$ (values $\mathcal{V}\in \mathbb{R}^{m\times d_v}$) via convolutional operators of $3\times 3$ kernels, where $d_q$ and $d_v$ are the feature dimensions of queries (keys) and values. To decouple the classification and regression tasks, $\mathcal{Q}$ and $\mathcal{K}$ are further projected into $\mathcal{Q}_{i}, \mathcal{K}_{i}, i\in\{sem,geo\}$ via individual MLP (implemented as 1D convolution).
To compute the weighted sum of the values $\mathcal{V}$ as the cross-view output $\mathcal{F}_i\in \mathbb{R}^{n\times d_v}$, the scaled dot-product is applied to obtain the cross-view attention weight $\mathcal{A}_i\in [0, 1]^{n\times m}$:
\begin{equation}
    \mathcal{A}_i = softmax(\frac{\mathcal{Q}_i\mathcal{K}_i^\mathcal{\top}}{\sqrt{d_q}}), i\in\{sem,geo\}     
\end{equation}
and the output will be $\mathcal{F}_i = \mathcal{A}_i\mathcal{V}$. The output $\mathcal{F}_i$ will be fed into individual Feed Forward Network $\mathcal{FFN}_i$ (FFN) to obtain the final results. We adopt the architecture widely used in previous works \cite{carion2020end,vaswani2017attention} as our FFN to ensure the non-linearity and diversity. The proposed VISTA is an one-stage method that directly generates proposals based on the features fused across views; such a design can leverage more information for accurate and efficient 3D detection.

\subsection{Decoupling Classification and Regression} \label{Decoupling}

The VISTA decouples the classification and regression tasks. After the shared convolutional operators, the queries and keys are further processed by individual linear projection to produce $\mathcal{Q}_{i}$ and $\mathcal{K}_{i}$, which will then participate in different attention modelling in terms of semantic information or geometric information. The motivation of such decoupling is the different impacts that the supervised signal of classification and regression will have on the training.

Given a query object in the scene, for classification, the attention module needs to aggregate the semantic cues from the objects in the global context to enrich the semantic information conveyed in the fused features. Such targets require the learned queries and keys to be aware of the commonalities among different objects of the same category, for the sense that the objects of the same category should match each other in regard to the semantic meaning. However, regression task can not take the same set of queries and keys since different objects have their own geometric characteristic (e.g. translation, scale, velocity, etc), the regression features should be diverse across different objects. Therefore, sharing the same queries and keys will induce conflicts to the attention learning during the joint training of classification and regression.

Furthermore, no matter single view or multi-view, the classification and regression results are all predicted from the same feature maps in the conventional voxel-based 3D detectors. However, due to the inherent property of the 3D scene, there exists inevitable occlusion and loss of texture information in the 3D point cloud, thus the 3D detector is difficult to extract the semantic features, leading great challenges to the learning of classification. On the contrary, the rich geometric information conveyed by the 3D point cloud relaxes the burden of understanding the geometric property of objects, which is the basis of learning regression task. As a result, during network training, there comes the imbalanced learning between the classification and regression, where the learning of classification is dominant by the regression. Such an imbalanced learning is a common issue in the 3D object detection involving classification and regression based on 3D point cloud, which will have negative impacts on the detection performances. To be concrete, the 3D detector will not be robust across different object categories (e.g. truck and bus) that have similar geometric features, as we shown in the Section \ref{Analysis}. 

To mitigate the issues described above, we are motivated to individually set up attention modelling for semantic and geometric information respectively. The output of the attention module are the $F_{sem}$ and $F_{geo}$ based on the constructed semantic and geometric attention weight. The supervision of classification and regression are applied on the $F_{sem}$ and $F_{geo}$ respectively, which guarantees the effective learning of the corresponding tasks.

\subsection{Attention Constraint} \label{Auxiliary Constrain}


The proposed VISTA faces many challenges when learns to model the cross-view correlation from the global context. The 3D scenes contain plenty of background points (approximately up to 95\%), and only a small portion are points of interest that contributes to the detection results. During the training of the cross-view attention, the massive background points will bring unexpected noise to the attention module. Moreover, the occlusion effect in the complex 3D scenes brings inevitable distortion to the attention learning. Consequently, the attention module tends to attend to the irrelevant regions, as shown in the Section \ref{Analysis}. The extreme case of the poor learning of attention is the global average pooling (GAP) operation, as we demonstrated in the Section \ref{sec:exp}, without any explicit supervisions, directly adopting the attention module for multi-view fusion yields performance similar to the GAP, which indicates that the attention module cannot model the cross-view correlations well.

To empower the attention module to focus at specific targets rather than generic points, we propose to apply a constraint on the variance of the learned attention weights. Thanks to the proposed constraint, we enable the network to have the ability to learn where to attend. By combining the attention variance constraint with the conventional classification and regression supervised signal, the attention module focuses at the meaningful targets in the scenes, as we shown in the Section \ref{Analysis}, thus producing high quality fused features. We formulate the proposed constraint as an auxiliary loss during training. For simplicity, we ignore the batch dimension, given a learned attention weight $\mathcal{A}\in\mathbb{R}^{N_{bev} \times N_{rv}}$ where the $N_{bev}$ and $N_{rv}$ are the number of pillars in BEV and RV respectively, the set of the scale and center location of ground-truth bounding boxes in x-y planes $\mathbb{B}=\{b_q|b_q=(w_q,h_q,x_q,y_q),q=1,...,N_{box}\}$, where the $N_{box}$ is the number of boxes in the scene. For each pillar in BEV, we calculate the real-world coordinates of its center based on the voxel size and obtain the set $\mathbb{C}=\{c_j|c_j=(x_j,y_j),j=1,...,N_{bev}\}$. The attention weights of each ground-truth bounding box are obtained by:

\begin{table*}[!tb]
\resizebox{\textwidth}{!}{%
\begin{tabular}{c|c|c|c|ccccccccccc}
\toprule
Methods      & NDS & mAP & runtime & car & truck & cons. & bus & trailer & barrier & motorcycle & bicycle & pedestrian & traffic cone \\ \midrule
PointPillars \cite{lang2019pointpillars} & 45.3 & 30.5 & 17ms & 68.4 & 23.0 & 4.1 & 28.2 & 23.4 & 38.9 & 27.4 & 1.1 & 59.7 & 30.8 \\
WYSIWYG \cite{hu2020you} & 41.9 & 35.0 & - & 79.1 & 30.4 & 7.1 & 46.6 & 40.1 & 34.7 & 18.2 & 0.1 & 65.0 & 28.8 \\
PointPainting \cite{vora2020pointpainting} & 59.2 & 46.4 & - & 77.9 & 35.8 & 15.8 & 36.2 & 37.3 & 60.2 & 41.5 & 24.1 & 73.3 & 62.4 \\
CBGS \cite{zhu2019class} & 63.3 & 52.8 & 55ms & 81.1 & 48.5 & 10.5 & 54.9 & 42.9 & 65.7 & 51.5 & 22.3 & 80.1 & 70.9 \\
CVCNet \cite{chen2020every} & 64.2 & 55.8 & 91ms & 82.7 & 46.1 & 20.7 & 45.8 & 46.7 & 69.9 & 61.3 & 34.3 & 81.0 & 69.7 \\ 
OHS \cite{chen2020object} & 66.0 & 59.3 & 60ms & 83.1 & 50.9 & 23.0 & 56.4 & 53.3 & 71.6 & 63.5 & 36.6 & 81.3 & 73.0 \\
CenterPoint \cite{yin2020center} & 67.3 & 60.3 & 70ms & \textbf{85.2} & 53.5 & 20.0 & 63.6 & \textbf{56.0} & 71.1 & 59.5 & 30.7 & \textbf{84.6} & 78.4 \\
\textbf{VISTA-OHS (Ours)}  & \textbf{69.8} & \textbf{63.0} & 69ms & 84.4 & \textbf{55.1} & \textbf{25.1} & \textbf{63.7} & 54.2 & \textbf{71.4} & \textbf{70.0} & \textbf{45.4} & 82.8 & \textbf{78.5} \\ \bottomrule
\end{tabular}%
}
\caption{3D detection results on the nuScenes test server. "cons." refers to construction vehicle.}
\label{Test Results}
\vspace{-0.4cm}
\end{table*}

\vspace{-0.4cm}
\begin{equation}
    \mathcal{A}_q = \mathcal{A}[p, :], \mathrm{s.t.} \left\{\begin{array}{cc}
        x_q - w_q/2 \leq x_p \leq x_q + w_q/2 &  \\
        y_q - h_q/2 \leq y_p \leq y_q + h_q/2 & 
    \end{array} \right.
\end{equation}
Then we formulate the variance constraint for all ground-truth bounding boxes as follows:

\begin{equation}
    \mathcal{L}_{var} = -\frac{1}{N_{box}}\sum_{q}^{N_{box}}\frac{1}{N_q}\sum_{i}^{N_q}Var(A_q[i])
\end{equation}
where $N_q$ is the number of pillars that enclosed by the $b_q$, Var($\cdot$) calculates the variance of the given vector.

\section{Implementation}

\textbf{Voxelization} We voxelize the point clouds according to the x,y,z axes. For nuScenes dataset, the range for voxelization are $[-51.2, 51.2]$m, $[-51.2,51.2]$m, and $[-5.0,3]$m in terms of x,y,z. For Waymo dataset, the range are $[-75.2,75.2]$m, $[-75.2,75.2]$m, and $[-2,4]$m. Unless specifically mentioned, all of our experiments are conducted in low voxelization resolution of $[0.1,0.1,0.1]$m of the x,y,z axes.

\textbf{Augmentation} The point clouds are randomly flipped according to the x,y axes, rotated around $z$ axis with a range of $[-0.3925,0.3925]$ rad, scaled with a factor ranging from 0.95 to 1.05, and translated with range $[0.2,0.2,0.2]$ m in x,y,z axes. The class-balanced grouping and sampling \cite{zhu2019class}, and the database sampling \cite{yan2018second} are adopted to increase the ratio of positive samples during training.

\textbf{Joint Training} We train the VISTA on various target assignment \cite{zhu2019class,chen2020object,yin2020center}. To train the network, the original loss for different target assignments is calculated, we recommend readers to refer to their original papers for more details of the loss. Briefly, we take classification and regression into account:
\begin{equation}
    \mathcal{L}_{target} = \lambda_{1}\mathcal{F}_{cls}(\hat{y},y) + \lambda_{2}\mathcal{F}_{reg}(\hat{b},b)
\end{equation}
where $\lambda_{1}$ and $\lambda_{2}$ are the loss weights, $\mathcal{F}_{cls}(\cdot,\cdot)$ is the classification loss function between ground-truth labels $\hat{y}$ and predictions $y$, $\mathcal{F}_{reg}(\cdot,\cdot)$ is the regression loss function between ground-truth bounding boxes $\hat{b}$ and predicted ones $b$.

The total loss $\mathcal{L}$ is the weighted sum of $\mathcal{L}_{target}$ and $\mathcal{L}_{var}$: $\mathcal{L}=\mathcal{L}_{target}+\lambda_{3}\mathcal{L}_{var}$. We set $\lambda_{1}$, $\lambda_{2}$, and $\lambda_{3}$ to 1.0, 0.25, 1.0. We apply Focal loss\cite{lin2017focal} as $\mathcal{F}_{cls}$, and L1 loss for $\mathcal{F}_{reg}$.

\section{Experiments} \label{sec:exp}

\begin{table}[!htb]
\resizebox{\linewidth}{!}{%
\begin{tabular}{cccccc|ccc}
\toprule
& Avg & Linear Atten & Conv Atten & Var Cons & Decouple & mAP & NDS & Runtime \\ \midrule
(a) & \multicolumn{1}{l}{} & \multicolumn{1}{l}{} & \multicolumn{1}{l}{} & \multicolumn{1}{l}{} &     & 59.5 & 66.0 & 60ms \\
(b)&\checkmark & & & & & 59.2 & 65.8 & 61ms\\
(c) & & \checkmark & & & & 58.7 & 65.9 & 63ms\\
(d) & & & \checkmark & & & 60.0 & 66.8 & 64ms\\
(e) & & & \checkmark & \checkmark & & 60.4 & 67.5 & 64ms\\
(f) &  & & \checkmark & \checkmark & \checkmark & \textbf{60.8} & \textbf{68.1} & 69ms\\ \bottomrule
\end{tabular}%
}
\caption{Ablation studies of VISTA on multi-view fusion. The performance is evaluated on the nuScenes validation set.}
\label{Ablation Study}
\vspace{-0.3cm}
\end{table}

\begin{table}[!htb]
\resizebox{\linewidth}{!}{
\begin{tabular}{c|cc|c|cc|c|cc}
\toprule
Method              & mAP  & NDS & Method              & mAP  & NDS & Method              & mAP  & NDS\\ \midrule
CBGS                & 51.9 & 61.5 & CenterPoint         & 56.4 & 64.8 & OHS                 & 59.5 & 66.0 \\ \midrule
V-CBGS        & \makecell[c]{\textbf{53.2}\\(+1.3)} & \makecell[c]{\textbf{62.8}\\(+1.3)} & V-CenterPoint & \makecell[c]{\textbf{57.6}\\(+1.2)} & \makecell[c]{\textbf{65.6}\\(+0.8)} & V-OHS         & \makecell[c]{\textbf{60.8}\\(+1.3)} & \makecell[c]{\textbf{68.1}\\(+2.1)} \\ \bottomrule
\end{tabular}%
}
\caption{3D detection results of VISTA-based state-of-the-art methods (V-method) on nuScenes validation set. For efficiency, all methods are experimented based on the low voxelization resolution configuration provided in their official codebase.}
\label{Performance Gain}
\vspace{-0.4cm}
\end{table}

\begin{table*}[!htb]
\resizebox{\textwidth}{!}{%
\begin{tabular}{ccccccccccccc}
                                             & \multicolumn{6}{c}{LEVEL\_1}                                                                                          & \multicolumn{6}{c}{LEVEL\_2}                                                                     \\ \toprule
\multicolumn{1}{c|}{\multirow{2}{*}{Method}} & \multicolumn{2}{c}{Vehicle}    & \multicolumn{2}{c}{Pedestrian} & \multicolumn{2}{c|}{Cyclist}                        & \multicolumn{2}{c}{Vehicle}    & \multicolumn{2}{c}{Pedestrian} & \multicolumn{2}{c}{Cyclist}    \\
\multicolumn{1}{c|}{}                        & \multicolumn{1}{c}{mAP} & mAPH & \multicolumn{1}{c}{mAP} & mAPH & \multicolumn{1}{c}{mAP} & \multicolumn{1}{c|}{mAPH} & \multicolumn{1}{c}{mAP} & mAPH & \multicolumn{1}{c}{mAP} & mAPH & \multicolumn{1}{c}{mAP} & mAPH \\ \midrule
\multicolumn{1}{c|}{StarNet \cite{ngiam2019starnet}}    & 61.5 & 61.0 & 67.8 & 59.9 & - & \multicolumn{1}{c|}{-}  & 54.9 & 54.5 & 61.1 & 54.0 & - & -    \\
\multicolumn{1}{c|}{PointPillars \cite{lang2019pointpillars}}            & 63.3                    & 62.8 & 62.1                    & 50.2 & 34.7                    & \multicolumn{1}{c|}{25.3} & 55.6                    & 55.1 & 55.9                    & 45.1 & 33.3                    & 24.3 \\
\multicolumn{1}{c|}{PPBA \cite{ngiam2019starnet}}                    & 67.5                    & 67.0 & 69.7                    & 61.7 & -                       & \multicolumn{1}{c|}{-}    & 59.6                    & 59.1 & 63.0                    & 55.8 & -                       & -    \\
\multicolumn{1}{c|}{RCD \cite{bewley2020range}}                     & 72.0                    & 71.6 & -                       & -    & -                       & \multicolumn{1}{c|}{-}    & 65.1                    & 64.7 & -                       & -    & -                       & -    \\
\multicolumn{1}{c|}{CenterPoint \cite{yin2020center}}             & 81.0                        & 80.6     & 80.5                        & 77.3     & 74.6                        & \multicolumn{1}{c|}{73.6}     &  73.4                       & 73.0     & 74.5                        & 71.5     & 72.1                        & 71.2     \\ \midrule
\multicolumn{1}{c|}{VISTA-CenterPoint}     &  \textbf{81.7}                       & \textbf{81.3}     &  \textbf{81.4}                       & \textbf{78.3}     & \textbf{74.9}                        & \multicolumn{1}{c|}{\textbf{73.9}}     & \textbf{74.4}                        & \textbf{74.0}     & \textbf{75.5}                        & \textbf{72.5}     & \textbf{72.5}                        & \textbf{71.6}      \\ \bottomrule
\end{tabular}%
}
\vspace{-0.2cm}
\caption{3D detection results on the Waymo test server}
\label{Waymo Results}
\vspace{-0.6cm}
\end{table*}

We evaluate VISTA on nuScenes dataset and Waymo Open Dataset. We test the efficacy of VISTA on three state-of-the-art methods with different target assignment: CBGS \cite{zhu2019class}, OHS \cite{chen2020object}, and CenterPoint \cite{yin2020center}. 

\subsection{Dataset and Technical Details}

\textbf{nuScenes Dataset} contains 700 scenes for training, 150 scenes for validation, and 150 scenes for testing. The dataset is annotated at 2Hz, in total 40000 key-frames are annotated with 10 object categories. Following \cite{zhu2019class}, we combine 10 sweeps for each annotated key-frame to increase the number of points. Average precision (mAP) and nuScenes detection score (NDS) are applied in our performance evaluation. NDS is a weighted average of mAP and other attributes metrics, including translation, scale, orientation, velocity, and other box attributes. During training, we follow CBGS \cite{zhu2019class} to optimize the model via Adam \cite{kingma2014adam} optimizer with one-cycle learning rate policy \cite{Gugger1cycle}. 

\textbf{Waymo Open Dataset} contains 798 sequences for training, 202 sequences for validation. Each sequence is of 20s duration and sampled at 10Hz with a 64 channels LiDAR, containing 6.1M vehicle, 2.8M pedestrian, and 67k cyclist boxes. We evaluate our networks on the metric of standard mAP and mAP weighted by heading accuracy (mAPH), which are based on the IoU threshold of 0.7 for vehicles, 0.5 for pedestrians and cyclist. The official evaluation protocol evaluates the methods in two difficulty levels: LEVEL\_1 for boxes with more than five LiDAR points, and LEVEL\_2 for boxes with at least one LiDAR point.

\subsection{Comparison with other methods}

We submitted the test results of the proposed VISTA-based OHS to the nuScenes test server. To benchmark the results, we follow \cite{yin2020center} to tune up the resolution for training and utilize the double flip for testing augmentation. Since our results are based on single model, methods which use ensemble models and extra data are not included in our comparisons. The test performance are shown in Table \ref{Test Results}. The proposed VISTA achieves state-of-the-art performance on nuScenes test set, outperforming all published methods in both overall mAP and NDS by large margins. Particularly, the performances on the motorcycle and the bicycle surpass the second best method CenterPoint \cite{yin2020center} by up to 48\% in mAP. Specifically, the performance gains in geometric-similar categories (e.g. truck, construction vehicle) confirm the efficacy of our proposed decoupling design.

To further validate the effectiveness of our proposed VISTA, we adopt the proposed VISTA to the CenterPoint \cite{yin2020center}, and submitted test results to the Waymo test server. During training and testing, we follow exactly the same rules as CenterPoint. The test performance is shown in Table \ref{Waymo Results}. VISTA brings significant improvements to CenterPoint on all categories of all levels, outperforming all the published results.

\subsection{Ablation Studies}

\textbf{VISTA in Multi-View Fusion} As shown in Table \ref{Ablation Study}, to demonstrate the superiority of the proposed VISTA, we conduct the ablation studies with OHS \cite{chen2020object} as our baseline (a) on the validation set of nuScenes dataset. As elaborated in the Section \ref{Auxiliary Constrain}, without attention constraint, the extreme case for learned attention weights will be global average pooling (GAP). To clarify, we manually obtain the RV features via GAP, and add them back to all BEV features as fusion. Such a GAP-based fusion method (b) drops the performance of baseline to 59.2\% in overall mAP, indicating the necessary of adaptively fusing multi-view features from global spatial context. Directly adopt the VISTA for multi-view fusion (d) results in 60.0\% in mAP. When replace the convolutional attention module to the conventional linear one (c), the overall mAP drops to 58.7\%, which reflects the significance of aggregating local cues for constructing cross-view attention. After adding the proposed attention variance constraint, as demonstrated (e), the performances raise to 60.4\% in overall mAP. The performance gains from (d) to (e) rows indicate that the attention mechanism can be well guided via the attention constraint, and as we will analyze in the Section \ref{Analysis}, the attention module is able to attend to regions of interest of whole scenes. Nevertheless, the shared attention modeling will bring the conflicts between the learning of classification and regression tasks, where the former task will further be dominant by the latter one in 3D object detection. As shown in (f), after decoupling the attention modeling, the performances raise from 60.4\% to 60.8\% in overall mAP, further verifying our assumption.

\textbf{VISTA in Different Target Assignments} The proposed VISTA is a plug-and-play multi-view fusion method and can be adopted in various recent advanced target assign strategies with slight modifications. To demonstrate the effectiveness and generalization ability of the proposed VISTA, we implement the VISTA on CenterPoint \cite{yin2020center}, OHS \cite{chen2020object}, and CBGS \cite{zhu2019class}, which are recent state-of-the-art methods. These methods stand for different main stream target assignments in terms of anchor-based or anchor-free manners. We evaluate the results on the validation set of the nuScenes dataset, for verification, all the methods are implemented based on the low voxelization resolution (i.e. $[0.1,0.1,0.1]$m of x,y,z axes) configuration provided by their official codebase\footnote{https://github.com/poodarchu/Det3D}\footnote{https://github.com/tianweiy/CenterPoint}. As demonstrated in Table \ref{Performance Gain}, all the three target assignments achieve large performance gains in both overall mAP and NDS scores (approximately 1.3\% and 1.4\% in mAP and NDS), indicating that the proposed VISTA can fuse multi-view features of universally high quality via dual cross-view attention mechanism.

\textbf{VISTA in Real-World Application} We demonstrate the runtime of the proposed VISTA on one RTX3090 GPU in Table \ref{Ablation Study}. Without any modifications, the baseline (a) runs at 60ms per frame. After adopting the convolutional attention module (d) in the baseline, the runtime increases to 64ms. We can observe from (e) and (f) that, while applying the proposed attention variance constraint does not influence the inference speed, the decoupling design costs 5ms, yet the extra decay is still negligible. Being running with such efficiency, we argue that the proposed VISTA completely meets the requirements of real-world application.

\subsection{Analysis of VISTA} \label{Analysis}

\begin{figure}[!tb]
    \centering
    \includegraphics[width=1.0\linewidth]{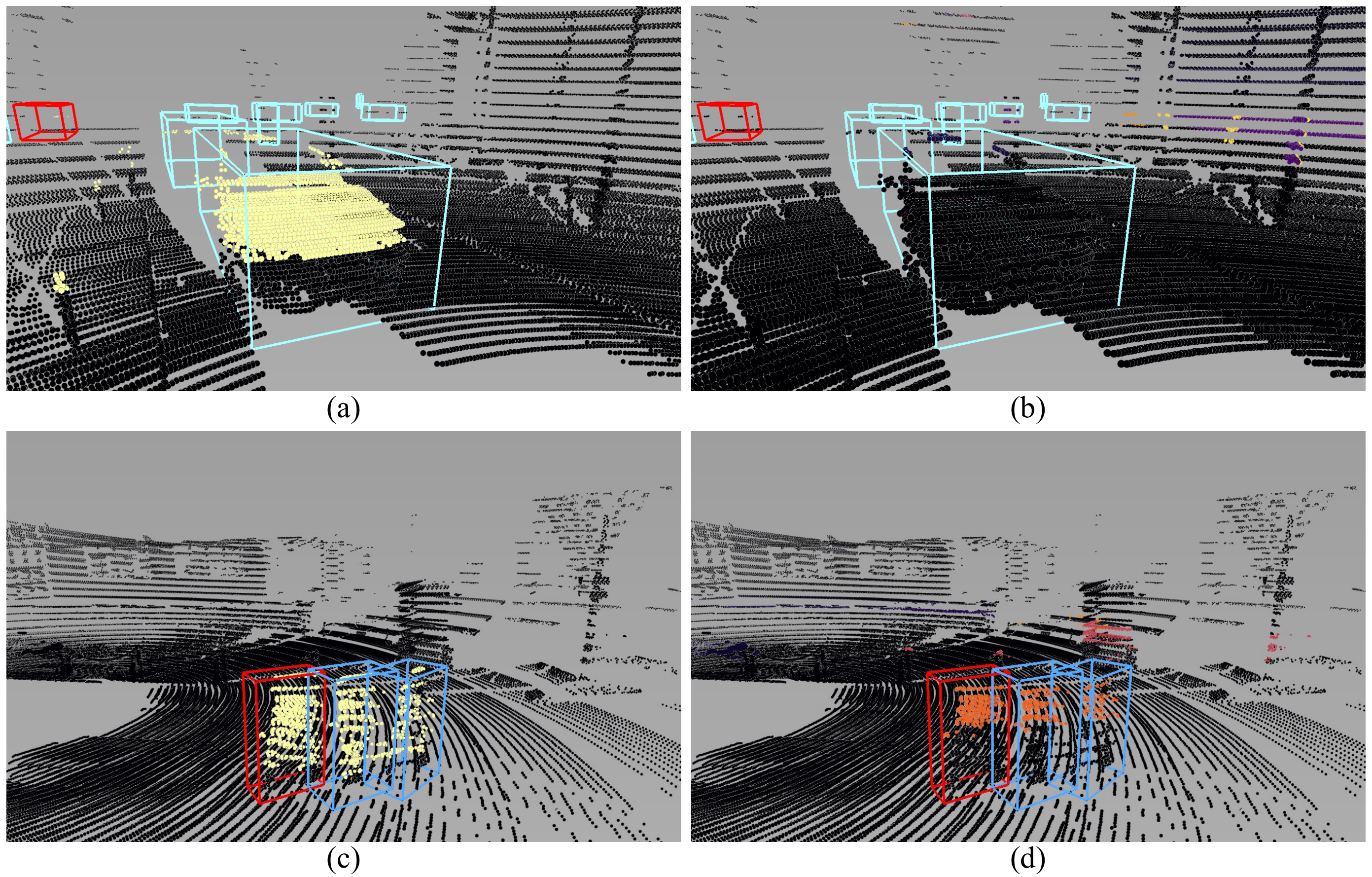}
    \caption{The visualization of learned attention map of VISTA with ((a) and (c)) and without ((b) and (d)) the attention variance constraint. Each row presents one sampled scene. The query bounding boxes are illustrated in red color. The brighter the color of the points, the higher the attention weight of the points.}
    \label{figAttenMap}
    \vspace{-0.5cm}
\end{figure}

We argue that the VISTA trained by the proposed attention constraint can capture the global and local correlations between BEV and RV, thus can effectively perform multi-view fusion for accurate bounding boxes prediction. To vividly present the effectiveness of attention variance constraint in training VISTA, we visualize the constructed cross-view correlations with and without attention variance constraint in Figure \ref{figAttenMap}. Given the area containing a bounding box from target view (BEV) to query the source view (RV), we get the corresponding cross-view attention weights for each pillar in the above area, and map the weight back to the origin point set for visualization. We observe that, without the proposed attention variance constraint, the learned attention weights hold small values for almost every pillars in the RV, resulting in an approximate global average pooling operation. In Figure \ref{figAttenMap} (b) and (d), the attention module attends to the background points far from the query cars and pedestrians, and the attention weights for each focusing region are relatively low. The attention module trained with attention variance constraint instead, highlights the objects with the same categories of the queries, as presented in Figure \ref{figAttenMap} (a) and (c). Especially, for the query cars, the attention module trained via attention variance constraint successfully attends on the other cars in the scenes. 

\begin{figure}[!tb]
    \centering
    \includegraphics[width=1.0\linewidth]{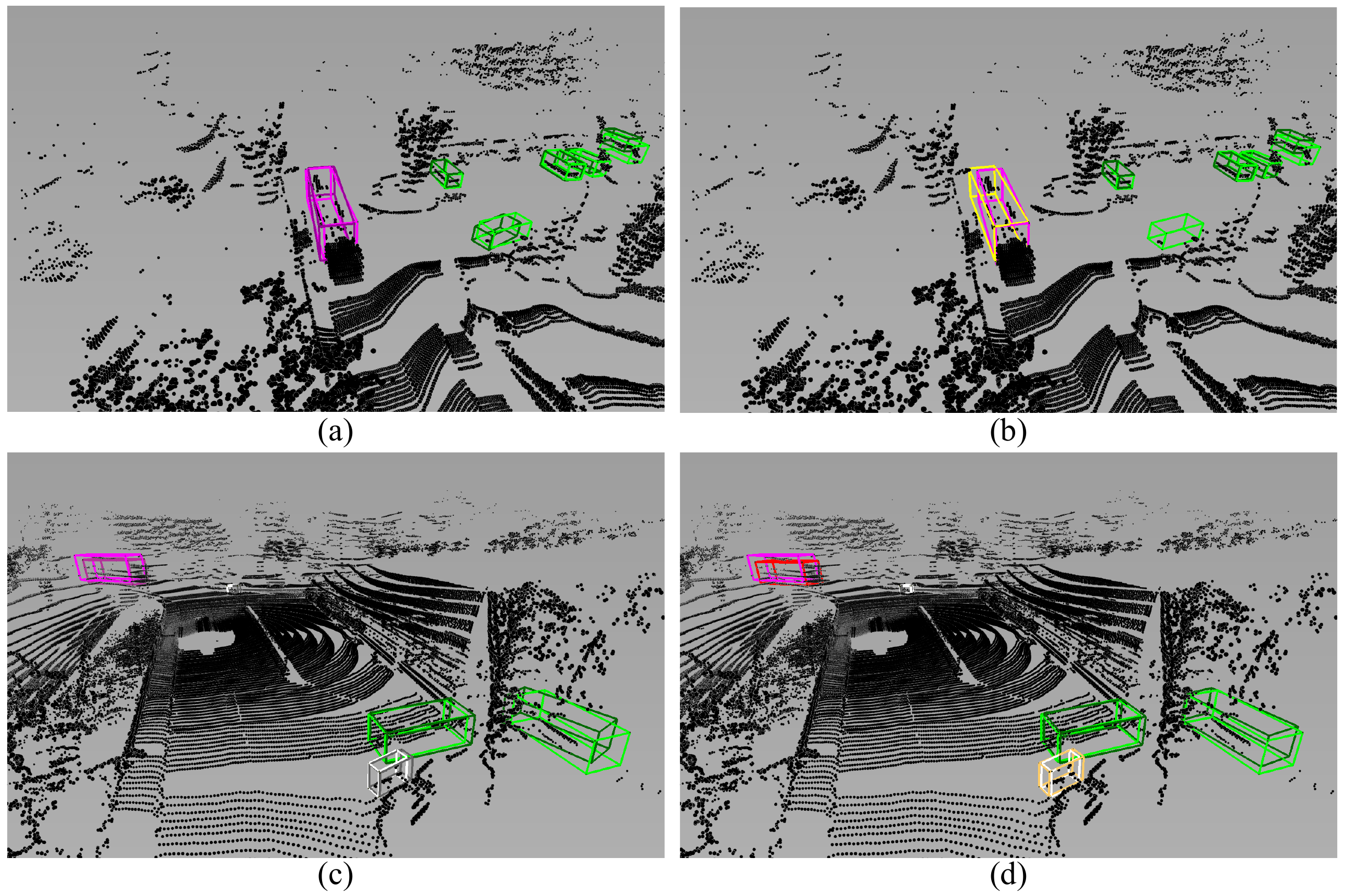}
    \caption{The visualization of the detection results learned with and without decoupling design. Each row represents one sample scene. The bounding boxes illustrated in light color refer to the ground truth bounding boxes, the boxes painted in dark color represent the correct prediction results, and the boxes illustrated in different accent color indicate the wrong predictions.}
    \label{figDecouple}
    \vspace{-0.5cm}
\end{figure}

The another key design of our proposed VISTA is the decoupling of the classification and regression tasks. The individual attention modeling for these two tasks mitigates the imbalanced learning issues, therefore the detection results are more accurate and reliable. To present the significance of our design, we present the detection results before and after the decoupling in the Figure \ref{figDecouple}. Each row represent one scene, and the left column presents the results with decoupling, the other column shows the results without decoupling. As illustrated in Figure \ref{figDecouple} (b) and (d), the 3D detector without decoupling design easily mistakes the objects A for the other B of similar geometric property, we term such phenomenon as A-to-B, such as bus (purple)-to-truck (yellow), bus (purple)-to-trailer (red), and bicycle (white)-to-motorcycle (orange), proving the imbalanced training of classification and regression tasks. Moreover, the confused predictions are not accurate when compare the right column to the left one. On the contrary, the VISTA with proposed decoupling design successfully distinguishes the categories of the objects, and predicts the tight bounding boxes, as shown in the Figure \ref{figDecouple} (a) and (c), demonstrating the efficacy of the proposed decoupling design.

\section{Discussion}

\noindent\textbf{Broader Impact} 3D object detection is vital to autonomous driving. Our proposed VISTA can identify the objects of interest precisely and comprehensively to ensure the safety in autonomous driving. However, misuse of technology will allow some malicious people or teams to invade and attack this perception part. Therefore, we encourage future research to mitigate these risks and make LiDAR sensor more robust and better.

\noindent\textbf{Limitations} We realize the multi-view fusion via the cross-view attention mechanism in the proposed VISTA. The attention mechanism requires large amount of data to train, and might perform poorly given insufficient data.

\noindent\textbf{Conclusion} In this paper, we propose VISTA, a novel plug-and-play multi-view fusion strategy for accurate 3D object detection. To empower the VISTA to have the ability to attend to the specific targets rather than generic points, we propose to constraint the variance of the learned attention weights. We decouple the classification and regression tasks to handle the issue of imbalanced training. Our proposed plug-and-play VISTA is able to produce high quality fused features for the prediction of proposals, and can be applied with various target assignment methods. Benchmarking on the nuScenes and Waymo datasets demonstrate the efficacy and generalization ability of our proposed method.

\noindent\textbf{Acknowledgement} This work is supported in part by the Guangdong R\&D key project of China (Grant No.: 2019B010155001) and the Program for Guangdong Introducing Innovative and Enterpreneurial Teams (Grant No.: 2017ZT07X183).

{\small
\bibliographystyle{ieee_fullname}
\bibliography{egbib}

\begin{thebibliography}{10}\itemsep=-1pt

\bibitem{bewley2020range}
Alex Bewley, Pei Sun, Thomas Mensink, Dragomir Anguelov, and Cristian
  Sminchisescu.
\newblock Range conditioned dilated convolutions for scale invariant 3d object
  detection.
\newblock {\em arXiv preprint arXiv:2005.09927}, 2020.

\bibitem{caesar2020nuscenes}
Holger Caesar, Varun Bankiti, Alex~H Lang, Sourabh Vora, Venice~Erin Liong,
  Qiang Xu, Anush Krishnan, Yu Pan, Giancarlo Baldan, and Oscar Beijbom.
\newblock nuscenes: A multimodal dataset for autonomous driving.
\newblock In {\em Proceedings of the IEEE Conference on Computer Vision and
  Pattern Recognition}, 2020.

\bibitem{carion2020end}
Nicolas Carion, Francisco Massa, Gabriel Synnaeve, Nicolas Usunier, Alexander
  Kirillov, and Sergey Zagoruyko.
\newblock End-to-end object detection with transformers.
\newblock In {\em European Conference on Computer Vision}, 2020.

\bibitem{chen2020mvlidarnet}
Ke Chen, Ryan Oldja, Nikolai Smolyanskiy, Stan Birchfield, Alexander Popov,
  David Wehr, Ibrahim Eden, and Joachim Pehserl.
\newblock Mvlidarnet: Real-time multi-class scene understanding for autonomous
  driving using multiple views.
\newblock In {\em International Conference on Intelligent Robots and Systems},
  2020.

\bibitem{chen2020every}
Qi Chen, Lin Sun, Ernest Cheung, and Alan~L Yuille.
\newblock Every view counts: Cross-view consistency in 3d object detection with
  hybrid-cylindrical-spherical voxelization.
\newblock {\em Advances in Neural Information Processing Systems}, 2020.

\bibitem{chen2020object}
Qi Chen, Lin Sun, Zhixin Wang, Kui Jia, and Alan Yuille.
\newblock Object as hotspots: An anchor-free 3d object detection approach via
  firing of hotspots.
\newblock In {\em European Conference on Computer Vision}, 2020.

\bibitem{chen2017multi}
Xiaozhi Chen, Huimin Ma, Ji Wan, Bo Li, and Tian Xia.
\newblock Multi-view 3d object detection network for autonomous driving.
\newblock In {\em Proceedings of the IEEE Conference on Computer Vision and
  Pattern Recognition}, 2017.

\bibitem{dosovitskiy2020image}
Alexey Dosovitskiy, Lucas Beyer, Alexander Kolesnikov, Dirk Weissenborn,
  Xiaohua Zhai, Thomas Unterthiner, Mostafa Dehghani, Matthias Minderer, Georg
  Heigold, Sylvain Gelly, et~al.
\newblock An image is worth 16x16 words: Transformers for image recognition at
  scale.
\newblock {\em arXiv preprint arXiv:2010.11929}, 2020.

\bibitem{graham2014spatially}
Benjamin Graham.
\newblock Spatially-sparse convolutional neural networks.
\newblock {\em arXiv preprint arXiv:1409.6070}, 2014.

\bibitem{Graham_2018_CVPR}
Benjamin Graham, Martin Engelcke, and Laurens van~der Maaten.
\newblock 3d semantic segmentation with submanifold sparse convolutional
  networks.
\newblock In {\em Proceedings of the IEEE Conference on Computer Vision and
  Pattern Recognition}, 2018.

\bibitem{graham2017submanifold}
Benjamin Graham and Laurens van~der Maaten.
\newblock Submanifold sparse convolutional networks.
\newblock {\em arXiv preprint arXiv:1706.01307}, 2017.

\bibitem{Gugger1cycle}
Sylvain Gugger.
\newblock The 1cycle policy.
\newblock {\em https://sgugger.github.io/the-1cycle-policy.html}, 2018.

\bibitem{guo2021pct}
Meng-Hao Guo, Jun-Xiong Cai, Zheng-Ning Liu, Tai-Jiang Mu, Ralph~R Martin, and
  Shi-Min Hu.
\newblock Pct: Point cloud transformer.
\newblock {\em Computational Visual Media}, 2021.

\bibitem{han2021transformer}
Kai Han, An Xiao, Enhua Wu, Jianyuan Guo, Chunjing Xu, and Yunhe Wang.
\newblock Transformer in transformer.
\newblock {\em arXiv preprint arXiv:2103.00112}, 2021.

\bibitem{hu2020you}
Peiyun Hu, Jason Ziglar, David Held, and Deva Ramanan.
\newblock What you see is what you get: Exploiting visibility for 3d object
  detection.
\newblock In {\em Proceedings of the IEEE Conference on Computer Vision and
  Pattern Recognition}, 2020.

\bibitem{kingma2014adam}
Diederik~P Kingma and Jimmy Ba.
\newblock Adam: A method for stochastic optimization.
\newblock {\em arXiv preprint arXiv:1412.6980}, 2014.

\bibitem{ku2018joint}
Jason Ku, Melissa Mozifian, Jungwook Lee, Ali Harakeh, and Steven~L Waslander.
\newblock Joint 3d proposal generation and object detection from view
  aggregation.
\newblock In {\em IEEE International Conference on Intelligent Robots and
  Systems}, 2018.

\bibitem{laddha2021mvfusenet}
Ankit Laddha, Shivam Gautam, Stefan Palombo, Shreyash Pandey, and Carlos
  Vallespi-Gonzalez.
\newblock Mvfusenet: Improving end-to-end object detection and motion
  forecasting through multi-view fusion of lidar data.
\newblock In {\em Proceedings of the IEEE Conference on Computer Vision and
  Pattern Recognition}, 2021.

\bibitem{lang2019pointpillars}
Alex~H Lang, Sourabh Vora, Holger Caesar, Lubing Zhou, Jiong Yang, and Oscar
  Beijbom.
\newblock Pointpillars: Fast encoders for object detection from point clouds.
\newblock In {\em Proceedings of the IEEE Conference on Computer Vision and
  Pattern Recognition}, 2019.

\bibitem{lin2017focal}
Tsung-Yi Lin, Priya Goyal, Ross Girshick, Kaiming He, and Piotr Doll{\'a}r.
\newblock Focal loss for dense object detection.
\newblock In {\em Proceedings of the IEEE International Conference on Computer
  Vision}, 2017.

\bibitem{liu2021swin}
Ze Liu, Yutong Lin, Yue Cao, Han Hu, Yixuan Wei, Zheng Zhang, Stephen Lin, and
  Baining Guo.
\newblock Swin transformer: Hierarchical vision transformer using shifted
  windows.
\newblock {\em arXiv preprint arXiv:2103.14030}, 2021.

\bibitem{meyer2019lasernet}
Gregory~P Meyer, Ankit Laddha, Eric Kee, Carlos Vallespi-Gonzalez, and Carl~K
  Wellington.
\newblock Lasernet: An efficient probabilistic 3d object detector for
  autonomous driving.
\newblock In {\em Proceedings of the IEEE Conference on Computer Vision and
  Pattern Recognition}, 2019.

\bibitem{ngiam2019starnet}
Jiquan Ngiam, Benjamin Caine, Wei Han, Brandon Yang, Yuning Chai, Pei Sun, Yin
  Zhou, Xi Yi, Ouais Alsharif, Patrick Nguyen, et~al.
\newblock Starnet: Targeted computation for object detection in point clouds.
\newblock {\em arXiv preprint arXiv:1908.11069}, 2019.

\bibitem{sheng2021improving}
Hualian Sheng, Sijia Cai, Yuan Liu, Bing Deng, Jianqiang Huang, Xian-Sheng Hua,
  and Min-Jian Zhao.
\newblock Improving 3d object detection with channel-wise transformer.
\newblock In {\em Proceedings of the IEEE International Conference on Computer
  Vision}, 2021.

\bibitem{sun2020scalability}
Pei Sun, Henrik Kretzschmar, Xerxes Dotiwalla, Aurelien Chouard, Vijaysai
  Patnaik, Paul Tsui, James Guo, Yin Zhou, Yuning Chai, Benjamin Caine, et~al.
\newblock Scalability in perception for autonomous driving: Waymo open dataset.
\newblock In {\em Proceedings of the IEEE Conference on Computer Vision and
  Pattern Recognition}, 2020.

\bibitem{vaswani2017attention}
Ashish Vaswani, Noam Shazeer, Niki Parmar, Jakob Uszkoreit, Llion Jones,
  Aidan~N Gomez, {\L}ukasz Kaiser, and Illia Polosukhin.
\newblock Attention is all you need.
\newblock {\em Advances in Neural Information Processing Systems}, 2017.

\bibitem{vora2020pointpainting}
Sourabh Vora, Alex~H Lang, Bassam Helou, and Oscar Beijbom.
\newblock Pointpainting: Sequential fusion for 3d object detection.
\newblock In {\em Proceedings of the IEEE Conference on Computer Vision and
  Pattern Recognition}, 2020.

\bibitem{yan2018second}
Yan Yan, Yuxing Mao, and Bo Li.
\newblock Second: Sparsely embedded convolutional detection.
\newblock {\em Sensors}, 2018.

\bibitem{yang2018pixor}
Bin Yang, Wenjie Luo, and Raquel Urtasun.
\newblock Pixor: Real-time 3d object detection from point clouds.
\newblock In {\em Proceedings of the IEEE Conference on Computer Vision and
  Pattern Recognition}, 2018.

\bibitem{yin2020center}
Tianwei Yin, Xingyi Zhou, and Philipp Kr{\"a}henb{\"u}hl.
\newblock Center-based 3d object detection and tracking.
\newblock {\em Proceedings of the IEEE Conference on Computer Vision and
  Pattern Recognition}, 2021.

\bibitem{zhou2020end}
Yin Zhou, Pei Sun, Yu Zhang, Dragomir Anguelov, Jiyang Gao, Tom Ouyang, James
  Guo, Jiquan Ngiam, and Vijay Vasudevan.
\newblock End-to-end multi-view fusion for 3d object detection in lidar point
  clouds.
\newblock In {\em Conference on Robot Learning}, 2020.

\bibitem{zhou2018voxelnet}
Yin Zhou and Oncel Tuzel.
\newblock Voxelnet: End-to-end learning for point cloud based 3d object
  detection.
\newblock In {\em Proceedings of the IEEE Conference on Computer Vision and
  Pattern Recognition}, 2018.

\bibitem{zhu2019class}
Benjin Zhu, Zhengkai Jiang, Xiangxin Zhou, Zeming Li, and Gang Yu.
\newblock Class-balanced grouping and sampling for point cloud 3d object
  detection.
\newblock {\em arXiv preprint arXiv:1908.09492}, 2019.

\bibitem{zhu2020deformable}
Xizhou Zhu, Weijie Su, Lewei Lu, Bin Li, Xiaogang Wang, and Jifeng Dai.
\newblock Deformable detr: Deformable transformers for end-to-end object
  detection.
\newblock {\em arXiv preprint arXiv:2010.04159}, 2020.

\end{thebibliography}
}

\clearpage
\textbf{\Large Appendix} \\
\appendix

\section{Implementation Details}

\begin{table*}[!htb]
\resizebox{\textwidth}{!}{%
\begin{tabular}{ccccc|c|cccccccccc}
\toprule
\multicolumn{1}{c}{Avg} & Linear Atten & Conv Atten & Var Cons & Decouple & mAP  & car  & truck & cons. & bus  & \multicolumn{1}{c}{trailer} & barrier & motorcycle & bicycle & pedestrian & traffic cone \\ \midrule
                        &              &            &          &                 & 59.5 & 84.2 & 56.5  & 19.7  & 65.6 & 36.0                        & 67.2    & 63.7       & 47.1    & 83.5       & 68.4         \\
\checkmark                     &              &            &          &          & 59.2 & 84.0 & 53.8  & 19.7  & 65.5 & 36.6                        & 64.5    & 67.4       & 48.2    & 83.1       & 67.1         \\
                      & \checkmark            &            &          &          & 58.7 & 84.6 & 52.4  & 18.4  & 65.5 & 36.0                        & 64.6    & 66.6       & 44.0    & 83.7       & 68.4         \\
                      &              & \checkmark          &          &          & 60.0 & 84.4 & 54.1  & 20.4  & 67.2 & 36.7                        & 64.4    & 66.6       & 50.6    & 83.5       & 69.6         \\
             &              & \checkmark          & \checkmark        &          & 60.4 & 84.7 & 55.8  & 19.8  & 67.1 & 36.2                        & 68.6    & 67.3       & 50.6    & 83.9       & 69.1         \\
    &              & \checkmark          & \checkmark        & \checkmark        & \textbf{60.8} & \textbf{84.8} & \textbf{57.2}  & \textbf{20.5}  & \textbf{67.6} & \textbf{36.8} & \textbf{69.0} & \textbf{67.7} & \textbf{50.7} & \textbf{84.1} & \textbf{69.7}         \\ \bottomrule
\end{tabular}%
}
\caption{The detailed ablation studies on the validation set of the nuScenes dataset. ``cons.'' refers to construction vehicle}
\label{Detailed Ablation}
\end{table*}

\begin{table*}[htb]
\resizebox{\textwidth}{!}{%
\begin{tabular}{c|c|cccccccccc}
\toprule
Method            & mAP           & car           & truck         & cons.         & bus           & trailer       & barrier       & motorcycle    & bicycle & pedestrian    & traffic cone  \\ \midrule
Decouple+Cls Var  & 60.6          & 84.8          & 57.0          & 20.4          & 66.9          & 36.2          & 68.3          & 65.7          & 49.1    & 83.5          & 69.5          \\ 
Decouple+Reg Var  & 60.5          & 84.7          & 55.4          & 19.3          & 66.4          & 36.0          & 68.6          & 67.1          & 50.2    & 84.0          & 69.3          \\ 
Decouple+Both Var & \textbf{60.8} & \textbf{84.8} & \textbf{57.2} & \textbf{20.5} & \textbf{67.6} & \textbf{36.8} & \textbf{69.0} & \textbf{67.7} & \textbf{50.7}    & \textbf{84.1} & \textbf{69.7} \\ \bottomrule
\end{tabular}%
}
\caption{Ablation studies of the attention variance constrain being applied on different tasks. ``Cls'' and ``Reg'' stand for the classification and regression, respectively. ``cons.'' refers to construction vehicle. The ablation studies are conducted on the validation set of the nuScenes dataset.}
\label{Variance Ablation}
\end{table*}

\noindent\textbf{Attention Module} Directly sequentializing the original feature maps into feature sequences will construct the intermediate attention maps with huge memory occupancy, which leads to unaffordable GPU memory costs. Considering efficiency and simplicity, the original feature maps are downsampled by average pooling before being passed into attention module, then the outputs of attention module are mapped back to original sizes via inverse mapping according to the pooling field. In practice, the kernel size of the average pooling is set to [4,4] and [4,1] for BEV and RV, respectively.

\noindent\textbf{Voxelization} We voxelize the point clouds according to the x,y,z axes. All ablation studies are conducted in low voxelization resolution of [0.1,0.1,0.1]m according to the x,y,z axes. To benchmark the results of our proposed VISTA-OHS on the nuScenes dataset, we follow the OHS \cite{chen2020object} to tune up the voxelization resolution to [0.08, 0.08, 0.08]m. In terms of the Waymo Open Dataset, following the official configurations provided by CenterPoint, we keep the low resolution unchanged.

\noindent\textbf{Training} We follow the CBGS \cite{zhu2019class} to train the proposed VISTA using Adam \cite{kingma2014adam} optimizer scheduled by one-cycle learning rate policy \cite{Gugger1cycle}. For Adam optimizer, the weight decay is set to 1e-2. And for the one-cycle learning rate policy, we set the max learning rate as 1e-3 for nuScenes and 3e-3 for Waymo Open Dataset, and the momentum is ranging from 0.95 to 0.85. We train the proposed VISTA on 4 RTX3090 GPUs for 20 epochs with batch size 16 on the nuScenes dataset, and for 36 epochs with batch size 16 on the Waymo Open Dataset. During training, the proposed attention variance constrain is applied on both the regression and classification branches of the decoupling architecture.

\noindent\textbf{Code} Our implementations are based on the open-sourced code released by CBGS \cite{zhu2019class} \footnote{https://github.com/poodarchu/Det3D} and CenterPoint \cite{yin2020center} \footnote{https://github.com/tianweiy/CenterPoint}. Code and experimental configurations will be released upon the acceptance of the paper.

\section{Extra Analysis of Decoupling Design}

\begin{figure}[!htb]
    \centering
    \includegraphics[width=1.0\linewidth]{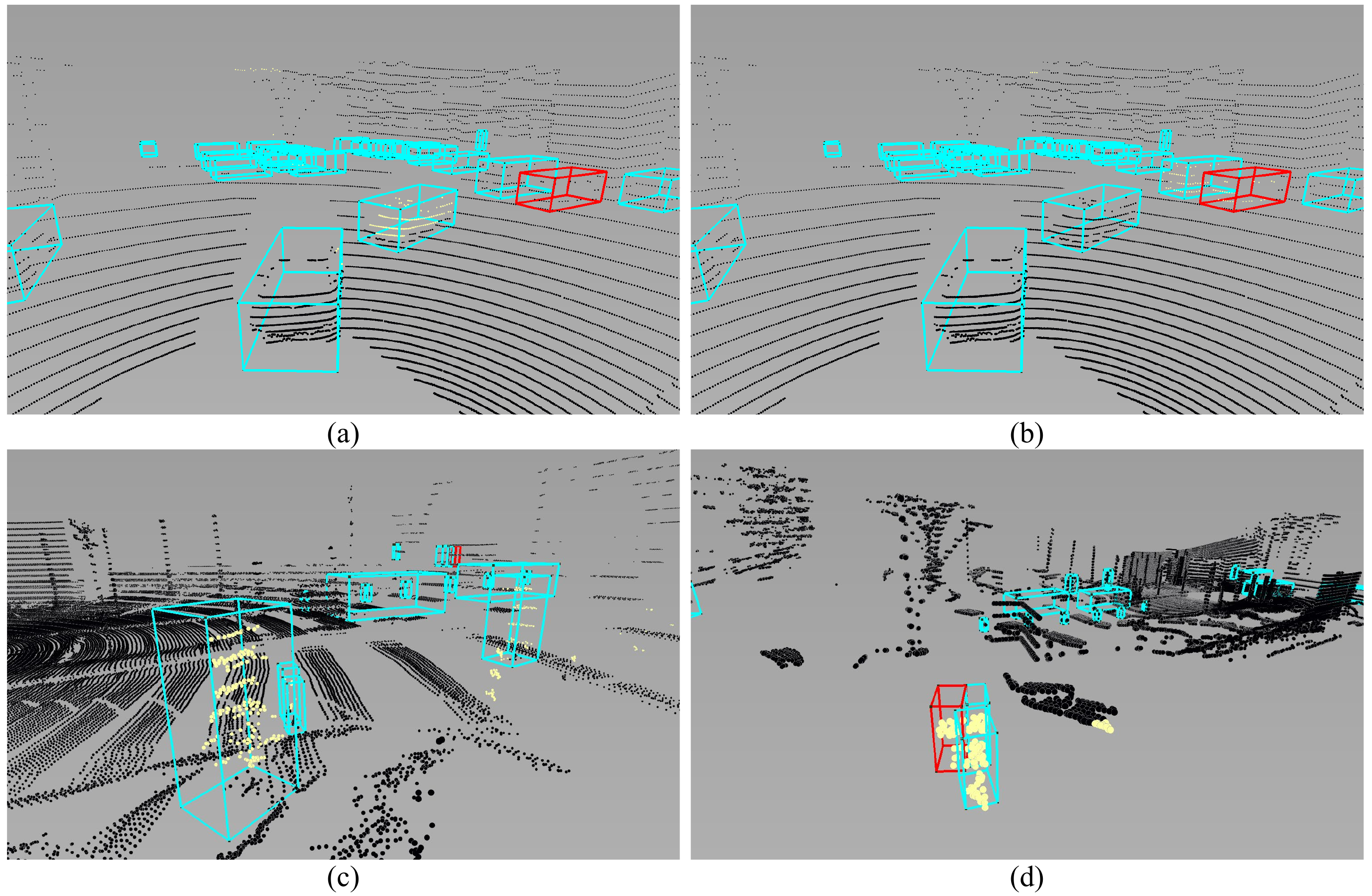}
    \caption{The attention weights learned by the classification and regression branches of the decoupled attention modeling. We choose two samples to present, each of which is demonstrated in one row. The left column illustrates the attention maps for classification tasks, the right column shows the regression one. The query boxes are painted in red, the brighter the points, the larger the attention weights.}
    \label{figDecouple_Supp}
\end{figure}

In this section, we present the attention weights in terms of regression and classification tasks in the Figure \ref{figDecouple_Supp}, which are learned in the decoupled attention modeling. Given the area containing a query bounding box from target view (BEV) to query the source view (RV), we get the corresponding cross-view attention weights for each pillar in the above area, and map the weight back to the origin point set for visualization. We observe that, different supervised signals lead to different attention weights. For classification task, the attention module tends to focus on the other objects in the whole scenes to enrich the semantic information contained in the fused features, as shown in the (a) and (c). To understand the geometric properties (e.g. scale, translation) of the query objects, the attention modeling for regression task instead, paying its attention to the local regions in which the query objects are, as we demonstrated in the (b) and (d). The different preferences of the individual attention modeling on the regions of interest further demonstrate the efficacy of our decoupling design.

During training, we apply the proposed attention variance constrain on both the classification and regression attention weights of the decoupled attention modeling. To further verify the different impacts that the classification and regression task will have on the network, we apply the attention variance constrain on different attention weights. The performances are demonstrated in the Table \ref{Variance Ablation}. We observe that, when apply the attention variance constrain on the attention weights of the classification task, the network yields better performances on the large objects (e.g. truck, construction vehicle). We argue that such performance gains are mainly due to the enriched semantic features. Since the most parts of the large objects in point cloud representation are empty, aggregating the corresponding dense cross-view features from the other objects is beneficial for the network to infer the categories of the objects. When it comes to the small objects (e.g. barrier, motorcycle, pedestrian), the small sizes of the objects make the network easier to consider the local context to understand the geometric properties, therefore, the regression task is better at handling the small objects when being applied the attention constrain. After adopting the proposed attention variance constrain in both classification and regression, the network benefits from the advantages on the large also the small objects, and yields the best performances, as shown in the last row of Table \ref{Variance Ablation}.

Nevertheless, the decoupling design definitely brings extra parameters. To further clarify that the performance gains come from the proposed decoupling structure, we conduct an experiment that uses a single attention and adds several convolutional layers to the detection head (Deeper Head); the setting keeps the number of parameters roughly the same. Validation results in Table \ref{Decoupling_Supp} show that the alternative setting of replacing the decoupling does not bring benefits, which further verifies the efficacy of our proposed decoupling design.
\begin{table}[ht]
\resizebox{\linewidth}{!}{%
\begin{tabular}{c|cc}
\hline
        & Deeper Head & Decoupling \\ \hline
mAP Gains  & 60.40$\rightarrow$60.45 (+0.05)   & 60.40$\rightarrow$60.81 (\textbf{+0.41}) \\ \hline
\end{tabular}%
}
\caption{The mAP gains on nuScenes Validation Set}
\label{Decoupling_Supp}
\end{table}

\section{Extra Analysis of Variance Constraint}

We apply the variance constraint on the positive samples during the training phase, which may form an ``upweight" of the positive samples. Hence, the stated performance gains benefit from the proposed variance constraint could be attributed to such an ``upweight" training. To clarify, we disable the variance loss and run a set of experiments that either scale up or scale down the background predictions, as shown in Table \ref{Variance}. Table \ref{Variance} shows that our variance loss is not equivalent to scaling of labels, thus verifying its efficacy.

\begin{table}[h]
\resizebox{\linewidth}{!}{%
\begin{tabular}{c|ccc}
\hline
        & Scaling down (x0.5) & Scaling up (x2) & Ours \\ \hline
mAP  & 59.6 & 60.4   & \textbf{60.8} \\ \hline
\end{tabular}%
}
\caption{The mAP on nuScenes Validation Set}
\label{Variance}
\end{table}

\section{Detailed Ablations}

In this section we extend the ablation studies to each category in Table \ref{Detailed Ablation} to present the category-wise performances. The ablation studies are conducted on the validation set of nuScenes dataset.

\end{document}